\documentclass{article}

\usepackage[preprint,nonatbib]{neurips_2024}

\usepackage[utf8]{inputenc} 
\usepackage[T1]{fontenc}    
\usepackage{hyperref}       
\usepackage{url}            
\usepackage{booktabs}       
\usepackage{amsfonts}       
\usepackage{nicefrac}       
\usepackage{microtype}      
\usepackage{xcolor}         
\usepackage{amsmath}
\usepackage{graphicx}

\newcommand{\ie}{\textit{i.e.}}
\newcommand{\eg}{\textit{e.g.}}

\newcommand{\etc}{\textit{etc.}}

\title{V-Express: Conditional Dropout for Progressive Training of Portrait Video Generation}

\author{
Cong Wang$^1$\thanks{Work done during an internship at Tencent AI Lab.}\ \ \thanks{Equal contribution.}\hspace{1em}
  Kuan Tian$^2$\footnotemark[2]\hspace{1em}
  Jun Zhang$^2$\thanks{Corresponding authors.}\hspace{1em}
  Yonghang Guan$^2$\hspace{1em}
  Feng Luo$^2$\hspace{1em}
  Fei Shen$^2$\\
  \textbf{Zhiwei Jiang}$^1$\footnotemark[3]\hspace{1em}
  \textbf{Qing Gu}\textsuperscript{1}\hspace{1em}
  \textbf{Xiao Han}$^2$\hspace{1em}
  \textbf{Wei Yang}$^2$\vspace{5px}\\
  $^1$ Nanjing University, $^2$ Tencent AI Lab\vspace{3px} \\
  \texttt{cw@smail.nju.edu.cn, \{kuantian, junejzhang\}@tencent.com,} \\
  \texttt{\{yohnguan, amandaaluo, ffeishen\}@tencent.com,} \\
  \texttt{\{jzw, guq\}@nju.edu.cn, \{haroldhan, willyang\}@tencent.com}\vspace{5px}\\
  \texttt{\url{https://tenvence.github.io/p/v-express/}}
}

\begin{document}

\maketitle

\begin{abstract}
In the field of portrait video generation, the use of single images to generate portrait videos has become increasingly prevalent. 
A common approach involves leveraging generative models to enhance adapters for controlled generation. 
However, control signals (\eg, text, audio, reference image, pose, depth map, \etc) can vary in strength.
Among these, weaker conditions often struggle to be effective due to interference from stronger conditions, posing a challenge in balancing these conditions. 
In our work on portrait video generation, we identified audio signals as particularly weak, often overshadowed by stronger signals such as facial pose and reference image. 
However, direct training with weak signals often leads to difficulties in convergence. 
To address this, we propose V-Express, a simple method that balances different control signals through the progressive training and the conditional dropout operation. 
Our method gradually enables effective control by weak conditions, thereby achieving generation capabilities that simultaneously take into account the facial pose, reference image, and audio. 
The experimental results demonstrate that our method can effectively generate portrait videos controlled by audio. 
Furthermore, a potential solution is provided for the simultaneous and effective use of conditions of varying strengths. 
\end{abstract}

\section{Introduction}

In recent years, diffusion models have dominated in the field of image generation \cite{dhariwal2021diffusion, ho2020denoising, peebles2023scalable, rombach2022high, sohl2015deep}.
This advancement has spurred the development of various methods and models aiming to enhance the quality and control of generated content \cite{zhang2023adding, mou2024t2i, ye2023ip}, especially in applications such as portrait video generation \cite{tian2024emo, xu2024vasa, wei2024aniportrait}.
The ability to generate high-quality portrait videos from single images is particularly valuable for numerous applications, including virtual avatars, digital entertainment, and personalized video content creation.

A common approach in portrait video generation involves leveraging generative models to enhance adapters for controlled generation. 
Control signals, such as text, audio, reference images, keypoints, and depth maps, play a crucial role in determining the quality and accuracy of the generated videos. 
However, these control signals can vary significantly in strength, leading to challenges in achieving balanced and effective control. 
In particular, weaker conditions, such as audio signals, often struggle to be effective due to interference from stronger conditions like keypoints and reference images. 
This imbalance poses a significant challenge, as the weaker signals are frequently overshadowed, making it difficult to achieve the desired level of control and synchronization.

To address these challenges, we propose V-Express, a novel method designed to balance different control signals through the progressive training and the conditional dropout operations. 
V-Express enables effective control by weaker conditions, ensuring that all signals contribute appropriately to the final output. 
Our method employs a Latent Diffusion Model (LDM) \cite{rombach2022high} to generate video frames, incorporating ReferenceNet, V-Kps Guider, and Audio Projection to handle various control inputs efficiently. 
The progressive training and conditional dropout strategy help mitigate the dominance of stronger signals, allowing weaker conditions, particularly audio, to have a more pronounced influence. This approach not only enhances the overall quality of the generated videos but also ensures better synchronization and control.

Our experimental results demonstrate that V-Express can effectively generate high-quality portrait videos with synchronized audio, maintaining consistency in facial identity and pose. 
The results highlight the potential of our method in providing a balanced approach to integrating multiple control signals of varying strengths.

\section{Method}

V-Express aims to generate a talking head video under the control of a reference image, an audio, and a sequence of V-Kps images. Among these conditions, the reference image controls the background and the face identity.
The audio guides the lip movement.
Each V-Kps image controls the facial position and pose of the corresponding frame of the output video.
A V-Kps image is crafted by mapping three facial keypoints (\ie, the left eye, the right eye, and the nose) onto an image set against a black background.
The lines with different colors connecting the left eye to the nose and the right eye to the nose form a `V' shape, which is the inspiration behind the term `V-Kps' image. It is well known that certainty between input and output makes the training process easier. Therefore, we believe that using the V-Kps image not only allows for excellent control over facial position and pose but also does not interfere with facial muscle movements and lip movements. 
Additionally, the movement of the eyes' keypoints in V-Kps images can control the eye blinking.

\subsection{Preliminaries}

In V-Express, we employ Latent Diffusion Model (LDM) to generate video frames. 
LDM executes both the diffusion and reverse processes of diffusion models \cite{ho2020denoising, sohl2015deep} within the latent space by Variational Auto-Encoder (VAE) \cite{kingma2013auto}. 
The input image $\mathbf{x}$ will be encoded by a VAE encoder $\mathcal{E}$ into a latent $\mathbf{z}=\mathcal{E}(\mathbf{x})$.
During the diffusion process, Gaussian noise is incrementally introduced into the latent $\mathbf{z}$, leading to its degradation into complete noise after $T$ steps. 
During the reverse process, the target latent $\mathbf{z}$ is generated from a sampled Gaussian noisy latent through iterative denoising using the diffusion model, and a VAE decoder $\mathcal{D}$ will decode the latent $\mathbf{z}$ into a output image $\mathbf{x}=\mathcal{D}(\mathbf{z})$.

When training the model, given a latent $\mathbf{z}_0=\mathcal{E}(\mathbf{x}_0)$ and the condition $\mathbf{c}$, the denoising loss is
\begin{equation}
    \mathcal{L}_\text{denoising} = \mathbb{E}_{
        \mathbf{z}_t, 
        \boldsymbol{\epsilon}\sim\mathcal{N}(\mathbf{0}, \mathbf{I}),
        \mathbf{c},
        t
    }
    \Vert
    \boldsymbol{\epsilon}_\theta(\mathbf{z}_t,\mathbf{c},t) - \boldsymbol{\epsilon}_t
    \Vert^2 \ .
\end{equation}
Among them, $\mathbf{z}_t=\sqrt{\alpha_t}\mathbf{z}_0+\sqrt{1-\alpha_t}\boldsymbol{\epsilon}_t$ is the noisy latent at timestep $t\in[1,T]$. 
$\boldsymbol{\epsilon}_t$ is the added noise.
$\boldsymbol{\epsilon}_\theta$ is the predicted noise by the diffusion model with parameters $\theta$.

V-Express uses Stable Diffusion v1.5\footnote{\url{https://huggingface.co/runwayml/stable-diffusion-v1-5}} (SDv1.5), which is a type of text-to-image LDM, as the backbone.
SDv1.5 is implemented based on U-Net \cite{ronneberger2015u}, and it contains two attention layers in each Transformer \cite{vaswani2017attention} block.
Both of them are a self-attention layer and a cross-attention layer, respectively.
The cross-attention layer aims to introduce the text CLIP \cite{radford2021learning} embeddings.

\subsection{Model Architecture}

\begin{figure}[t]
    \centering
    \includegraphics[width=1\textwidth]{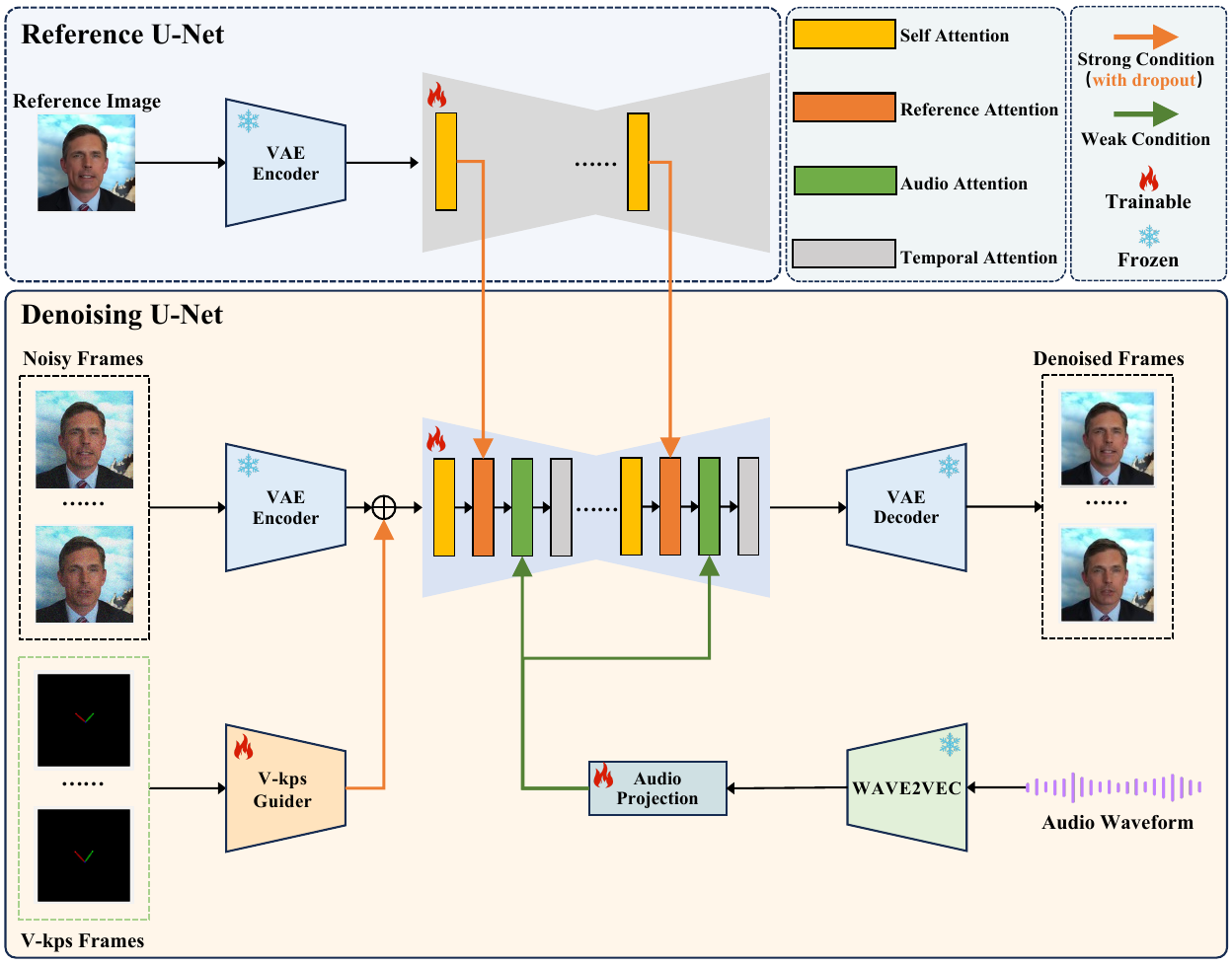}
    \caption{The framework of V-Express.}
    \label{fig:framework}
\end{figure}

As shown in Figure~\ref{fig:framework}, the backbone of V-Express is a denoising U-Net, which denoises the input multi-frame noisy latent under the conditions.
The architecture of the denoising U-Net closely mirrors that of SDv1.5, with a key difference being the presence of four attention layers in each Transformer block, instead of two.
The first attention layer is a self-attention layer, just like in SDv1.5. 
The second and third attention layers are cross-attention layers. 
The second attention layer, known as the Reference Attention Layer, encodes the relationship with the reference images. 
The third attention layer, called the Audio Attention Layer, encodes the relationship with the audio. 
These three attention layers are all spatial-wise attention layers. 
Finally, the fourth attention layer, termed the Motion Attention Layer, is a temporal-wise self-attention layer that captures the temporal relationships between the video frames.
Additionally, V-Express incorporates three crucial modules: \textbf{ReferenceNet}, \textbf{V-Kps Guider}, and \textbf{Audio Projection}, which are utilized to encode the reference image, the V-Kps images, and the audio, respectively.

\paragraph{ReferenceNet} 

In V-Express, the reference image provides crucial guidance to the generation process due to the strong consistency required for facial identity and background.
To achieve this, we use a ReferenceNet to encode the reference images.
The ReferenceNet shares the same architecture as SDv1.5 and is aligned in parallel with the denoising U-Net.
Within each Transformer block of ReferenceNet, the reference features are extracted by the self-attention layers. 
These reference features then serve as the key and value in the second attention layer of the corresponding Transformer block within the denoising U-Net.
It is important to note that ReferenceNet is only used to encode the reference image. 
Therefore, noise is not added to the reference image, and only a single forward pass is conducted during the diffusion process. In addition, empty text is fed into the cross-attention layer of ReferenceNet in order not to introduce other additional information.

\paragraph{V-Kps Guider}

Each V-Kps image has a strong spatial correspondence with the corresponding target frame. 
To leverage this, we use a V-Kps Guider to encode the V-Kps images. 
The V-Kps Guider is a lightweight convolutional model that encodes each V-Kps image into a V-Kps feature, which matches the shape of the latent.
Subsequently, before being fed into the denoising U-Net, the multi-frame latents are directly added with the encoded V-Kps features.

\paragraph{Audio Projection}

To synchronize the lip movements in the output video with the corresponding audio, we preprocess the audio to align it with each video frame. 
First, we resample the audio at a sampling rate of 16 kHz and then use a pretrained audio encoder, Wav2Vec2 \cite{baevski2020wav2vec}, to convert the resampled audio into an embedding sequence. 
Typically, the number of these embeddings does not exceed twice the number of video frames.
To ensure an even alignment between the audio embeddings and the video frames, we linearly interpolate the original audio embeddings into a sequence that is twice the length of the video. 
This results in each video frame being matched with two audio embeddings.
Furthermore, the current lip status may be affected not only by the current audio segment but also by the audio segments immediately before and after it. 
To capture this contextual information, we concatenate the audio embeddings of the two preceding and two subsequent audio segments with the current audio embeddings, forming the final set of audio embeddings.
Before inputting these audio embeddings into the denoising U-Net, we apply Audio Projection using a Q-Former \cite{li2023blip}.

\paragraph{Motion Attention Layer}

To generate consecutive frames, it is crucial to consider the temporal relationships between video frames.
V-Express addresses this by employing additional Motion Attention Layers to encode these temporal dynamics.
Specifically, given a hidden state $\mathbf{h} \in \mathbb{R}^{b\times f \times d \times h \times w}$, where $b$, $f$, $d$, $h$, and $w$ represents the batch size, the number of frames, the feature dimension, the height, and the width, the Motion Attention Layers effectively capture the inter-frame dependencies.
This is achieved by reshaping the hidden state $\mathbf{h} \in \mathbb{R}^{(b\times h \times w) \times f \times d}$ and performing the self-attention along the frame sequence.
The Motion Attention Layers are trained to learn motion patterns, ensuring that the generated frames exhibit smooth and coherent transitions, maintaining temporal consistency throughout the video sequence.

\subsection{Progressive Training Strategy}

The training pipeline of V-Express is progressive, which contains three stages.

\textbf{Stage I} focuses on single-frame generation, where V-Express receives a one-frame input and computes the corresponding denoising loss. 
During this stage, only ReferenceNet, V-Kps Guider, and the denoising U-Net are trained. 
Note that among the four attention layers, the weights of the Audio Attention Layers and Motion Attention Layers are not updated. 
These two attention layers are skipped due to the zero-initialized "to-out" linear layers.

\textbf{Stage II} focuses on multi-frame generation, where V-Express receives multi-frame inputs and predicts the added multi-frames noise. 
During this stage, only the Audio Projection, Audio Attention Layers, and Motion Attention Layers are trained, while the parameters of the other modules remain fixed.

\textbf{Stage III}, similar to Stage II, also focuses on multi-frames generation. 
However, this stage involves global fine-tuning, where all parameters are updated.

\subsection{Training Tricks}

\paragraph{Mouth Loss Weight}

The mouth do not dominate the entire image. 
To accelerate the convergence of lip synchronization, we assign a larger weight to the denoising loss for the mouth area.

\paragraph{Conditional Dropout}

In our experiments, we found that the model tends to learn a shortcut pattern, directly copying the V-Kps-affected reference image to the generated frames. 
This likely occurs because the control signals from the V-Kps and reference images are much stronger than those from the audio. 
This imbalance causes the weaker condition (e.g., audio) to struggle to be effective due to interference from the stronger condition (e.g., V-Kps and reference image).
To balance these conditions with varying control strengths, we employ conditional dropout to disrupt this shortcut. 
Specifically, in stages II and III, we randomly zero out the reference features and V-Kps features for some frames. 
This ensures that the generation of these frames is not directly controlled by the reference and V-Kps features. 
Instead, they rely on guidance from the Motion Attention Layers, effectively mitigating the shortcut problem.

\subsection{Inference}

During inference, given a audio with $t$ seconds and a predefined video FPS $f$, the number of generated frames is $[tf]$.
The provided V-Kps sequence is then linearly interpolated to match this length.
The frames are generated in multiple segments, with adjacent segments having some overlapping frames.
The latents of these overlapping frames are simply averaged before decoding.

When the V-Kps does not match the corresponding keypoints of the reference image, a retargeting method is needed. 
We provide a naive retargeting method.
First, we select the V-Kps frame that most closely resembles the reference image by comparing the distance ratios between the left eye to the nose and the right eye to the nose. 
Next, we compute rescaling parameters based on the face size ratio.
Finally, the rescaled V-Kps is adjusted according to the original nose trace.

\section{Experiments}

\subsection{Implementation}

V-Express is trained by HDTF~\cite{zhang2021flow}, VFHQ~\cite{xie2022vfhq}, and other collected videos.
The facial regions in these videos are cropped and resized to 512×512. 
The total training dataset comprises approximately 300 hours of video. 
Since VFHQ contains videos without audio, it is only used to train Stage I.

V-Express is implemented using PyTorch \cite{paszke2019pytorch} and optimized with Adam \cite{kingma2014adam}. 
During the multi-frame training in Stage II and Stage III, the number of frames is set to 12. 
For inference, the length of video segments is also 12, with 4 overlapping frames between segments.
The mouth loss weight is set as 100.
The dropout rate of the V-Kps images and the reference images are set as 50\% and 20\%, respectively.

\subsection{Quantitative Comparison}

We simply compare V-Express with 2 methods, which are Wav2Lip \cite{prajwal2020lip} and DiffusedHeads \cite{stypulkowski2024diffused}.

We evaluate all the methods on the subsets of two public datasets, both of which are TalkingHead-1KH \cite{wang2021one} and AVSpeech \cite{ephrat2018looking}, respectively.
For each dataset, 100 videos are randomly sampled.
Note that the videos of these two datasets are not included into the training data.

We use five metrics to quantitative evaluation.
Fréchet Image Distance (FID)~\cite{heusel2017gans} and Fréchet Video Distance (FVD)~\cite{unterthiner2018towards} evaluate the quality of the generated frames \cite{ye2024real3d} and the generated videos, respectively. 
To assess the preservation of facial identity between the reference image and the generated video, we extract their facial embeddings and compute the cosine similarity between them. 
$\Delta$FaceSim represents the difference between the average face similarity of the ground-truth frames with the reference image and that of the generated frames with the reference image. 
KpsDis evaluates the alignment of the facial poses in the generated frames with that of the target V-Kps by computing the distance between the corresponding V-Kps. 
Finally, we employ the SyncNet \cite{chung2017out} score to evaluate the lip synchronization with the audio.

As shown in Table~\ref{tab:quantitative-comparison}, although V-Express does not achieve the best lip synchronization, it excels in video quality and alignment with other control signals.

\begin{table}[t] \scriptsize \setlength{\tabcolsep}{4.0pt} 
    \centering
    \caption{Quantitative comparison.}
    \label{tab:quantitative-comparison}
    \begin{tabular}{lcccccccccc}
    \toprule
    & \multicolumn{5}{c}{TalkingHead-1KH} & \multicolumn{5}{c}{AVSpeech} \\ \cmidrule(lr){2-6} \cmidrule(lr){7-11}
    & FID & FVD & $\Delta$FaceSim & KpsDis & SyncNet & FID & FVD & $\Delta$FaceSim & KpsDis & SyncNet \\ \midrule
    
    Ground-Truth 
    & -- & -- & 0.00 (80.69) & -- & 4.961 & -- & -- & 0.00 (84.56) & -- & 5.132 \\ \midrule
    
    Wav2Lip~\cite{prajwal2020lip} 
    & 29.06 & 250.95 & 10.77 (91.46) & 41.60 & \textbf{6.890} & 26.71 & 251.40 & 7.28 (91.84) & 42.93 & \textbf{6.623} \\
    
    DiffusedHeads~\cite{stypulkowski2024diffused} 
    & 115.41 & 344.69 & / & / & / & 104.61 & 363.10 & / & / & / \\
    
    \textbf{V-Express (Ours)} 
    & \textbf{25.81} & \textbf{135.82} & \textbf{3.63} (84.32) & \textbf{3.28} & 3.480 & \textbf{23.38} & \textbf{117.93} & \textbf{0.28} (84.84) & \textbf{2.78} & 3.793 \\
    \bottomrule
    \end{tabular}
\end{table}

\subsection{Results}

\begin{figure}[t]
    \centering
    \includegraphics[width=1\textwidth]{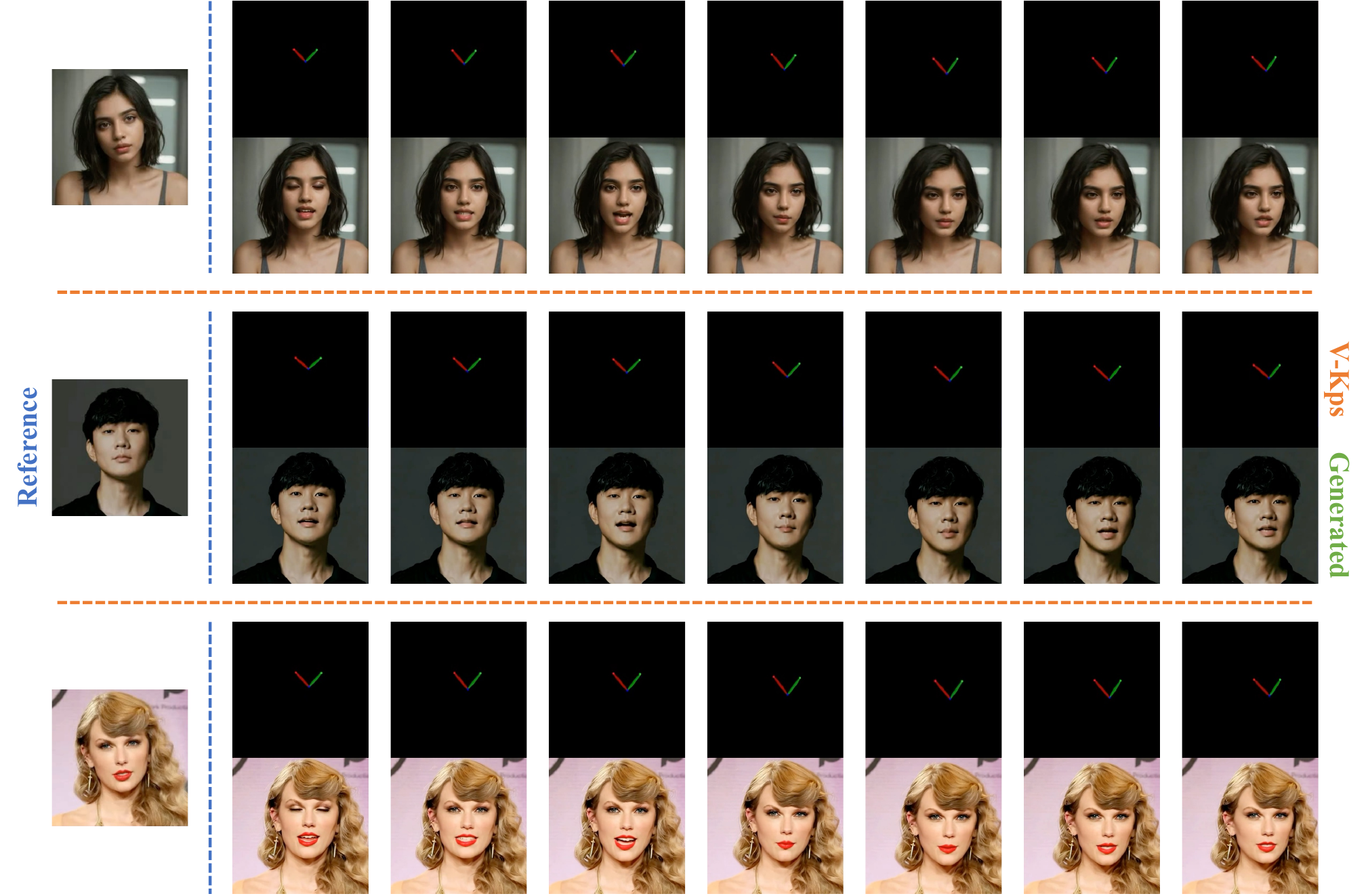}
    \caption{Some results of V-Express.}
    \label{fig:results}
\end{figure}

\begin{figure}[t]
    \centering
    \includegraphics[width=1\textwidth]{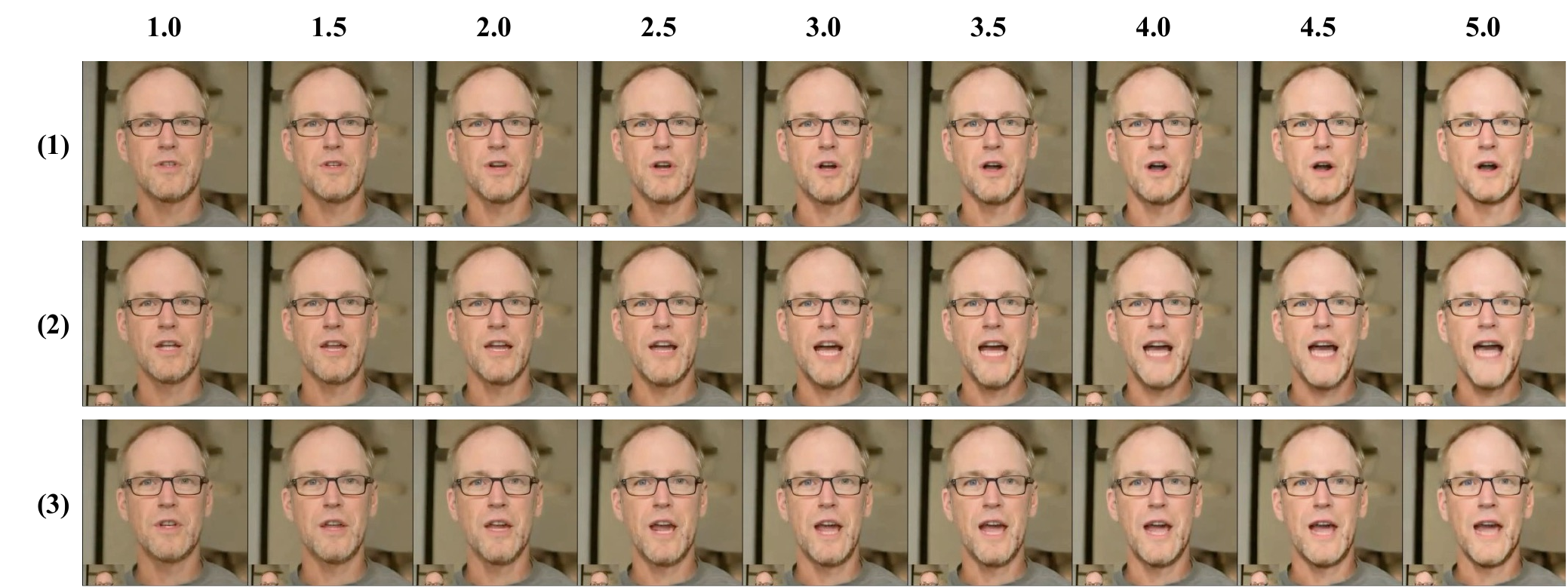}
    \caption{Effect of audio attention weights.}
    \label{fig:attn-weights}
\end{figure}

As shown in Figure~\ref{fig:results}, V-Express effectively generates portrait videos controlled by audio and V-Kps. 
Additionally, we find that the weight of the cross-attention hidden states can vary the strength of the corresponding control signal.
As illustrated in Figure~\ref{fig:attn-weights}, a larger audio attention weight results in more pronounced mouth movements.
To reduce the influence of the reference image, decreasing the weight of the reference attention will be effective.

\section{Conclusion}

Our proposed V-Express for portrait video generation addresses the challenge of balancing control signals of varying strengths. 
By employing progressive training and conditional dropout operations, we have enabled effective control by weaker conditions, such as audio signals, while maintaining the influence of stronger signals like facial pose and reference images. 
Our experimental results show that V-Express can generate high-quality portrait videos synchronized with audio inputs. 
This method not only enhances the overall quality of the generated videos but also provides a solution for the simultaneous and effective use of diverse control signals, paving the way for more advanced and balanced portrait video generation systems.

\paragraph{Future Work}

Although V-Express can effectively generate high-quality portrait videos controlled by V-Kps and audio, there are still some drawbacks that we aim to address in future work.
(1)~\textbf{Multilingual Support}: The current version performs better with English and does not work as well with other languages, such as Chinese. 
This may be because that the Wav2Vec model we use does not support multiple languages.
Integrating a multilingual audio encoder could potentially resolve this limitation.
(2) \textbf{Reduce Computational Burden}: The current version has a slow generation speed due to the autoregressive diffusion process for multi-frame generation. 
A possible solution is to use LCM \cite{luo2023latent} or LCM-LoRA \cite{luo2023lcm} to reduce the number of inference denoising steps.
(3) \textbf{Explicit Face Attribute Control}: The facial attributions cannot be explicitly controlled by user-given signals. 
Using an annotated dataset to train a new cross-attention layer may be a possible solution.

{
\small
\bibliographystyle{plain}
\bibliography{neurips_2024}
}

\end{document}